\documentclass[submission, copyright]{eptcs}
\providecommand{\event}{FMAS 2023} 

\usepackage{iftex}
\usepackage{amsmath,amssymb,amsfonts}
\usepackage{textcomp}
\usepackage{algorithmic}
\usepackage{graphicx}
\usepackage{cite}

\ifpdf
\else
\fi

\title{Runtime Verification of Learning Properties for Reinforcement Learning Algorithms}
\author{Tommaso Mannucci \qquad\qquad Julio de Oliveira Filho
\institute{Intelligent Autonomous Systems}
\institute{TNO -- Netherlands Organisation for Applied Scientific Research \\
The Hague, The Netherlands}
\email{tommaso.mannucci@tno.nl \qquad\qquad julio.deoliveirafilho@tno.nl}
}
\def\titlerunning{Runtime verification of learning properties for reinforcement learning algorithms}
\def\authorrunning{T. Mannucci, J. de Oliveira Filho}

\begin{document}
\maketitle

\begin{abstract}

Reinforcement learning (RL) algorithms interact with their environment in a trial-and-error fashion. Such interactions can be expensive, inefficient, and timely when learning on a physical system rather than in a simulation. This work develops new runtime verification techniques to predict when the learning phase has not met or will not meet qualitative and timely expectations. This paper presents three verification properties concerning the quality and timeliness of learning in RL algorithms. With each property, we propose design steps for monitoring and assessing the properties during the system's operation.

\end{abstract}
\section{Introduction}
\label{introduction}

Reinforcement learning (RL) \cite{Sutton} is a bio-inspired approach to machine learning which formalizes the notion of “trial-and-error” and "learn-by-doing". RL enables systems to learn during operation based on sequential interactions with the environment. During their learning phase, RL algorithms encourage decisions that led to good results in the past while avoiding detrimental choices.  This simple concept is at the base of some stunning results in robotics automation\cite{Singh2021}, natural language processing\cite{Li2016}, and computerised gaming, such as the Atari\cite{Mnih}, StarCraft \cite{Vinyals} video games, and the ancient tabletop games of Chess, Shogi, and Go \cite{Silver}. 


Due to the runtime and interactive nature of RL algorithms, there has been an increasing demand for guarantees about their learning; e.g., that it will be concluded within a certain amount of time or interactions when done in an operational environment. It is also necessary to guarantee the agent learned its solution space well enough during the learning phase. Offering such guarantees for RL algorithms is a challenging task.  Traditional testing is often not possible due to the difficulty of acquiring a representative set of operating conditions\cite{Kenton}.  Formal testing methods do not yet scale well, and many RL algorithms use “black box” components\cite{Ehlers}, such as artificial neural networks.  The underlying models for these algorithms are uninformative and highly dimensional; and the individual effect of their many parameters on the overall performance is not apparent nor easy to assess. 

This work proposes new \textit{Runtime Verification}(RV) techniques for checking properties of the learning phase of RL algorithms. RV is an engineering discipline concerned with checking a system behaviour during its execution\cite{Bartocci}.  Many RL algorithms' properties require such a runtime verification approach.  Safety, timeliness, and robustness properties, for example, can become invalid when RL algorithms engage in learning during operation. This happens because properties observed in the system during design time might no longer hold as the system changes by learning from new data. Timeliness properties, such as the duration of the learning phase, depend on the order and variety of interactions presented to the RL agent. 

The specific contributions of this paper are:
\begin{itemize}
\item We propose formal specifications for three verification properties related to the learning phase of RL algorithms:
\begin{description}

\item[Quality of learning] This property measures \emph{how well has the agent learned its environment}.  It is related to the variety and frequency of experiences presented to an agent during the learning phase.
\item[Distance to optimal policy] This property assesses \emph{how far the current learned policy is from the optimal policy}.  
\item[Time to learn] A property that estimates the \emph{the amount of interactions the learning process will need to evaluate a (new) policy}. 
\end{description}
\item Along with each property, we propose the design of an RV monitor able to assess the property from observations and during the learning phase.  We discuss which information should be observable from the learning phase, how to collect it systematically, and how to use observations to assess the properties.  We propose ideas for the monitor's implementation and how it can be efficiently instrumented in an RL-based system.
\end{itemize}

The paper is organized as follows. Section \ref{background} provides a short review of related and relevant work. Section \ref{fundamentals} introduce basic concepts of RL and RV we need to derive the verification properties. Section \ref{properties} derives formal specifications for the properties, proposes monitoring techniques and examples. Section \ref{conclusions} discusses our conclusions and further work.
\section{Related Work}
\label{background}
Verification of RL safety properties has received the main priority in the literature because RL algorithms obviously need to explore in a safe way\cite{Corsi, Zhu, Pathak} if they are to learn in real-world setups. Pathak et al.\cite{Pathak} and Zhu et al. \cite{Zhu} go beyond system checks and specify how the result of their verification procedure can be used to enforce the safety constraints after a system re-design. Other specific RL frameworks \cite{Mason, Hunt, Anderson} use monitors to guide the system preventing the agent from violating the properties specified. Safety is an important aspect and has received significant attention in research. In this work however, we focus on other two important runtime properties of RL algorithms: learning quality and timeliness.
 
Verification of properties for quality of learning has received less attention than safety properties in research. There are guarantees of convergence for specific algorithms, such as Q-learning\cite{Watkins1992} and SARSA\cite{Singh}.  But this only means that such an algorithm will eventually learn. We differ in which we provide explicit ways to assess how much a system already learned after a set of experiences. Like us, Van Wesel and Goodloe \cite{Wesel} propose off-line and online verification techniques for quality-of-learning properties. However, their approach does not leverage from knowledge of the inner structure of the algorithm. Xin et al.\cite{Xin} introduce the concept of \textit{exploration entropy} to guide the learning until the final policy is of sufficient quality; their approach differs from proper verification in that it steers, and thus interferes with, the learning process.

Verification of timeliness of the learning phase is even less prominent.  Szepesvari \cite{Szepesvari} investigates the rate of convergence of Q-learning, and Potapov and Ali \cite{Potapov} analyze the influence of learning parameters on the convergence speed. But none of them provide an approach to verify if an RL algorithm will be able to learn within a desired number of interactions. This work differs from the aforementioned previous work by (1) providing formal specifications for quality and timeliness of the RL learning process and (2) providing monitoring techniques to check such properties at runtime. The monitors we propose do not modify the behaviour of the learning phase.

This work drives from the analysis of Markov decision processes in Mannor et al. \cite{Mannor}.  We use their approach on the calculation of estimates for the RL value function and its bias and variance estimates. We extend many of their results to define formal verification properties.  And we show how to monitor the RL algorithm during the learning phase to check these properties on-line.
\section{Fundamentals}
\label{fundamentals}

\subsection{Reinforcement learning}
Reinforcement Learning is a class of machine learning (ML) algorithms that solves control and decision problems.  This work targets RL variants which can be modelled as finite Markov decision processes (MDP) such as Q-learning\cite{Watkins1992}.  We use the finite MDP problem structure to formally derive verification properties, and later, to design the verification monitor.  An MDP problem is defined by a tuple $\{S, A, T, R, \gamma \}$, where $S$ is a set possible of environment states. $A$ is a set of actions an agent can take at each state and $T := S \times A \times S \rightarrow [0,1]$ is a probabilistic state transition function. $R := S \times A \times S \rightarrow \mathbb{R}$ is a reward function attributing a payoff for each state transition and action.  $\gamma \in [0, 1)$ is a discount factor over past rewards. 
 
During the execution of the RL algorithm, an agent operates in a sequence of distinct steps.  During the $n^{th}$ step, the agent observes the current state $s_n \in S$, chooses and performs an action $a \in A$.  The action is chosen according to a (probabilistic) policy $\pi := S \times A \rightarrow [0,1]$ .  This causes the environment to transition to a subsequent state $s_{n+1} \in S$ according to $T$. For its action and the new state achieved, the agent receives an instantaneous reward $r$ according to $R$. Rewards obtained from state $s$ following policy $\pi$ are accrued into the so-called value function: 

\begin{equation}
\label{value_definition}
    V^{\pi}(s) := E[\sum_{n=0}^N  \gamma^n  R(s_n, \pi(s_n),T(s_n, a))]
\end{equation}

\noindent where $\gamma^n$ is the $n^{th}$ power of $\gamma$, and $s_n$ is the $n^{th}$ state encountered after starting in $s=s_0$. A policy is optimal (indicated as $\pi^*$) if it maximizes $V^{\pi^*}(s)$ for all states. Note that the expectation operator in Eq.~\ref{value_definition} is due to the potential stochasticity of $\pi$ and $T$. In many cases, it is convenient to define the action value function

\begin{equation}
    Q^{\pi}(s,a) = E[R(s,a, T(s, a)) + \gamma V^\pi(T(s, a))]
\end{equation}

\noindent indicating the value obtainable in $s$ by taking action $a$ and following the policy thereafter. 

Temporal difference(TD)\cite{Sutton} is a method to solve RL problems when functions $T$ and $R$ are unknown. In this method, the value function is randomly initialized and a policy $\pi$ is followed. The reward observed at every transition is used to correct the value function, with a chosen learning rate $\alpha$ dictating the speed of correction. TD learning is proven to converge to a fixed value function $V^\pi$, given the agent has had enough and representative interactions with the environment.  In section \ref{properties}, we will use this fact as an intuitive notion for the quality of learning.

To define RV properties and monitors, we will use two results from Mannor et al. \cite{Mannor}.  First, estimates $\hat{T}$ and $\hat{R}$ (of transition and reward functions $T$ and $R$, respectively) can be reconstructed from observing transitions during the learning phase.  As a consequence, it is also possible to produce a value function estimate $\hat{V}^\pi$ directly via Eq.~\ref{value_definition}. Second, estimators for the bias and variance of the this value function estimate can be obtained as follows.  Under the assumption that all state action combinations are visited at least once:

\begin{equation}
\label{bias}
    bias(\hat{V}^\pi) =  \gamma^2 XQV^\pi + \gamma XB + o(\frac{1}{\mathrm{min}_{s,a}N(s,a)}) \approx  \gamma^2 XQ\hat{V}^\pi + \gamma XB;   
\end{equation}

\begin{equation}
\label{covariance}
   cov(\hat{V}^\pi) = XWX^T + o(\frac{1}{\mathrm{min}_{s,a}N(s,a)}) \approx XWX^T
\end{equation}

\noindent where $X$, $Q$, $W$ and $B$ are matrices computed from transitions and from $\hat{T}$ and $\hat{R}$.  The derivation and interpretation of these matrices is out of scope for this paper; the interested reader is referred to \cite{Mannor}. That being said, this result provides the bias and variance of the value function, which reflect the uncertainty of the agent due to lack of data. This is confirmed by the fact that $ bias(\hat{V}^\pi)$ and $ cov(\hat{V}^\pi)$ tend to zero by construction\cite{Mannor} if $\mathrm{min}_{s,a}N(s,a) \rightarrow \infty$. Thus estimates  $\hat{T}$, $\hat{R}$ and  $\hat{V}^\pi$ will converge to their true value with more transitions.

%

%
The value function $V^\pi$ can be used to iteratively improve the agent policy via the so-called \textit{policy improvement}:
 
\begin{equation}
\label{policy_improvement}
    \pi_{k+1}(s) =  \mathrm{argmax}_a E[R(s,a, T(s,a)) + V^{\pi_k} (T(s,a))]= \mathrm{argmax}_a Q^{\pi_k}(s,a)
\end{equation}

\noindent which converges during a proper learning experience to the optimal policy $\pi^*$ yielding the optimal value function $V^{\pi^*}$.
 
Replacing $ V^{\pi_k}$ with $ \hat{V}^{\pi_k}$ will yield an estimate $\hat{V}^{\pi^*}$ of the optimal value function $V^{\pi^*}$. However, such an estimate will be biased, as optimization will favor actions for which the expected cumulative reward is overestimated. \cite{Mannor} recognizes the problem and proposes to divide the set of all transitions into a calibration set and a validation set to mitigate this inconvenience. With this, compute calibrated estimates $\hat{T}_\mathrm{cal}$, $\hat{R}_\mathrm{cal}$ and $\hat{V}^{\pi_{cal}}$, and apply policy improvement to obtain $\pi^*_\mathrm{cal}$. From the validation set, obtain the transition and reward function estimates $\hat{T}_\mathrm{val}$ and $\hat{R}_\mathrm{val}$. Finally, compute $\hat{V}^{\pi_{cal}}$  via Eq.~\ref{value_definition}.

\subsection{Runtime verification}

Runtime Verification(RV) is an engineering discipline that combines (semi-)formal methods and monitoring of the system operation to check if a system's behaviour conforms to requirements.  \textit{Monitors} assess the system based on carefully collected observations of the system behaviour -- called \textit{traces}.  Traces must conform to formally specified \textit{properties}.      Therefore, Bartocci et al. \cite{Bartocci} indicate three steps to define an RV technique:
\begin{enumerate}
\item First, it is necessary to describe the property under verification using an unambiguous specification.  Mathematical or logical formulations which can be assessed on system traces are the most common.  

\item Second, it is necessary to design a monitor, which is a component able to collect and to assess traces of the system.  Assessment here means any analysis steps necessary to evaluate the trace against the specified property. 

\item Third, it is necessary to \textit{instrument} the system.  That is, insert observation mechanisms for correctly collecting the system traces.  Good instrumentation minimally interferes with the system behaviour and performance.    
\end{enumerate}

This work follows these three steps for each of the proposed properties. For each property, we derive a property specification, and provide monitoring steps to observe the system and calculate the property.  The monitor can be implemented to assess all the three properties concurrently and based on the same observed traces.  

\subsection{Use case: police patrol scheduling}
\label{use_case_example}

We sketch a fictional but typical example for an RL-based learning system.  In the remainder of the paper, we will use this example to illustrate the defined properties and discuss aspects relevant to the RV monitors and instrumentation.

A police department wants to use a new scheduling system for police night shifts in a city with frequent crime. This system will use an RL module that learns the most effective patrol schedules between three risk areas: the docks, the slums, and the bus station. Specifically, every hour between 00:00 am and 06:00 am, a patrol car is assigned to one of the three areas, for a total of six shifts per night. The RL algorithm for such a system has a state set $S := \{ (t, loc) | loc \in \{ \text{docks}, \text{slums}, \text{station} \}, \: t \in [0,5] \}$ and an action set $A := \{ \text{docks}, \text{slums}, \text{station} \}$.  

For our approach, the underlying problem structure ($S$ and $A$) must be known to the designer of the RV monitor. Neither the system nor the monitor designer knows the transition function $T$ and the reward function $R$ to be used. The transition function is not known because when a patrol car is sent to a location, it may take more or less time to complete the patrol. As a consequence it may miss a shift or terminate a shift early. The reward function cannot be estimated in advance as it is unknown which criminal activities can be prevented and where. However, it is decided to assign a reward between 0 and 3 in proportion to the severity of the spotted criminal activity, with 0 corresponding to no crime and 3 corresponding to a very severe crime.

\section{Runtime Verification for quality and timeliness of RL Learning}
\label{properties}

In this section, we propose formal specifications for three verification properties related to the learning phase of RL-algorithms: \emph{quality of learning}, \emph{distance to optimal policy}, and \emph{time to learn}.

\subsection{How well has the agent learned its environment?}
\label{quality_property}

The first question is how to estimate if the agent has learned ``enough'' from its environment. Intuitively, a TD learning agent has learned enough if the current value function $V^\pi(s)$ is close to its converged value (for all states). This choice is justified by the fact that the value function is related to both environmental stochastic functions $T$ and $R$, as well as to the fact that correctly estimating $V^\pi$ means correctly estimating the performance of the agent's policy as well \footnote{Assuming that $\pi$ and $T$ act ergodically concerning $S$ and $A$, i.e., that all state-action combinations are visited with non-zero probability.}. Unfortunately, due to the stochasticity of both policy and environment, analizing the value function error in time can lead to premature convergence assessments. Instead, we propose using bias and covariance of the value function estimate $\hat{V}$. Since these reflect the lack of gathered data of the agent, they can be used to assess when enough transitions have been accumulated by the agent to learn from, even if the agent does not make direct use of the estimates $\hat{T}$ and $\hat{R}$ , but relies on another method to solve the MDP problem, e.g., TD learning.
The procedure, based on Eq.~\ref{bias} and Eq.~\ref{covariance}, is as follows:

\begin{enumerate}
    \item query the policy of the system $\pi$;  
    \item read traces, assumed in the form $\{s, a, r, s'\}$, i.e., the MDP transitions;
    \item compute off-policy estimates $\hat{T}$ and $\hat{R}$ based on observed transitions, as well as on-policy estimates $\hat{T}^\pi (s, s'):= \hat{T}(s, \pi(s),s')$ and $\hat{R}^\pi(s):=\hat{ R}(s, \pi(s), \hat{T}(s, \pi(s))$;
    \item compute value $\hat{V}^\pi$ via Eq.~\ref{value_definition}.
    \item compute matrices $X$, $W$, $B$ and $Q$ from $\hat{T}$ and $\hat{R}$;
    \item compute bias and covariance, ignoring in first approximation the $o(\frac{1}{\mathrm{min}_{s, a}N(s, a)})$ term;
    \item compute relative bias $bias_{rel} := bias(\hat{V}^\pi(s))/\hat{V}^\pi(s)$ and relative variance  $\sigma_{rel}:=\sigma(\hat{V}^\pi(s))/\hat{V}^\pi(s)$;
    \item compare relative bias and variance with predefined upper thresholds; if for all states the two are below their respective thresholds, the property is satisfied.
\end{enumerate}

The procedure is straightforward to follow and implement but presents a few limitations as well. First, the monitor must have the memory to store past traces to be able to recompute the estimates $\hat{T}^\pi$ and $\hat{R}^\pi$. Second, the monitor must have access to the policy of the system. This is different than observing a signal trace in that the policy is not a signal, but a function utilized internally by the system. In case the policy is not observable within the system, it would be recommendable to use an estimate $\hat{\pi}$ from the observed actions in the trace. In this case, however, some error in the estimated bias and covariance can be expected given that the policy is an estimate in itself. Third, the procedure described here is not incremental, i.e., it does not provide for a method to update the estimates of bias and covariance at time $k+1$ given the trace at time $k+1$ and previous estimates of bias and covariance at time $k$; however, the calculations can be repeated by storing the previous traces. Finally, the method as presented requires that all state-action combinations are visited at least once (i.e., $\min_{s, a} N(s, a)>0$). If this condition is not verified, then the bias and covariance cannot be computed.

\subsubsection*{Example}

Looking back at the example sketched in Sec.\ref{use_case_example}, imagine the RL scheduling system has been provided with an initial exploratory policy assumed to perform decently, so as not to waste the patrolling effort while the RL system gathers information.
How long should information be gathered with this policy? To answer this question, a monitor is designed, following the given procedure, to verify the property

\begin{equation}
\label{property_1}
   \max_s \, bias_{rel}(s) < 0.05 \bigwedge  \max_s \, \sigma_{rel}(s) < 0.02
\end{equation}

\noindent which indicates that the bias and variance are small, respectively $2\%$ and $5\%$ of the estimated value, and therefore the epistemic uncertainty on the value function is low (note that these thresholds are for illustration purposes). The monitor shall inform whether this property is violated, or unverified\footnote{It might appear unsound that the monitor shall be able to report that the property is unverified since the inequality formulation of the property can in theory be always verified. However, at the start of the exploration, the condition $\min_{s, a} N(s, a)>0$ will not be verified for the applicability of the method. Therefore, the monitor will produce a ``property unverified'' response.}, or satisfied at each moment. Note that the monitor will not be able to predict when the property will be satisfied, it can only say if it is so at the current time. Furthermore, the fact that the property is initially unsatisfied does not mean that it cannot be satisfied eventually. 

After a sufficient amount of traces is collected, so that $\min_{s, a} N(s, a)>0$, the initial estimates of bias and variance can be generated. However, $\hat{T}$ and $\hat{R}$ will initially be poor estimates of $T$ and $R$, so the corresponding bias and variance are likely to be outside of the ranges provided by Eq. \ref{property_1}. Therefore, the monitor will produce a ``property violated'' result. The longer the trace, however, the more $\hat{T}$ and $\hat{R}$ will resemble the true matrices. Accordingly, bias and variance will reduce, until eventually Eq. \ref{property_1} will be true. The monitor will then produce a ``property satisfied'' response. 

\begin{figure}[t]
\includegraphics[width=17cm]{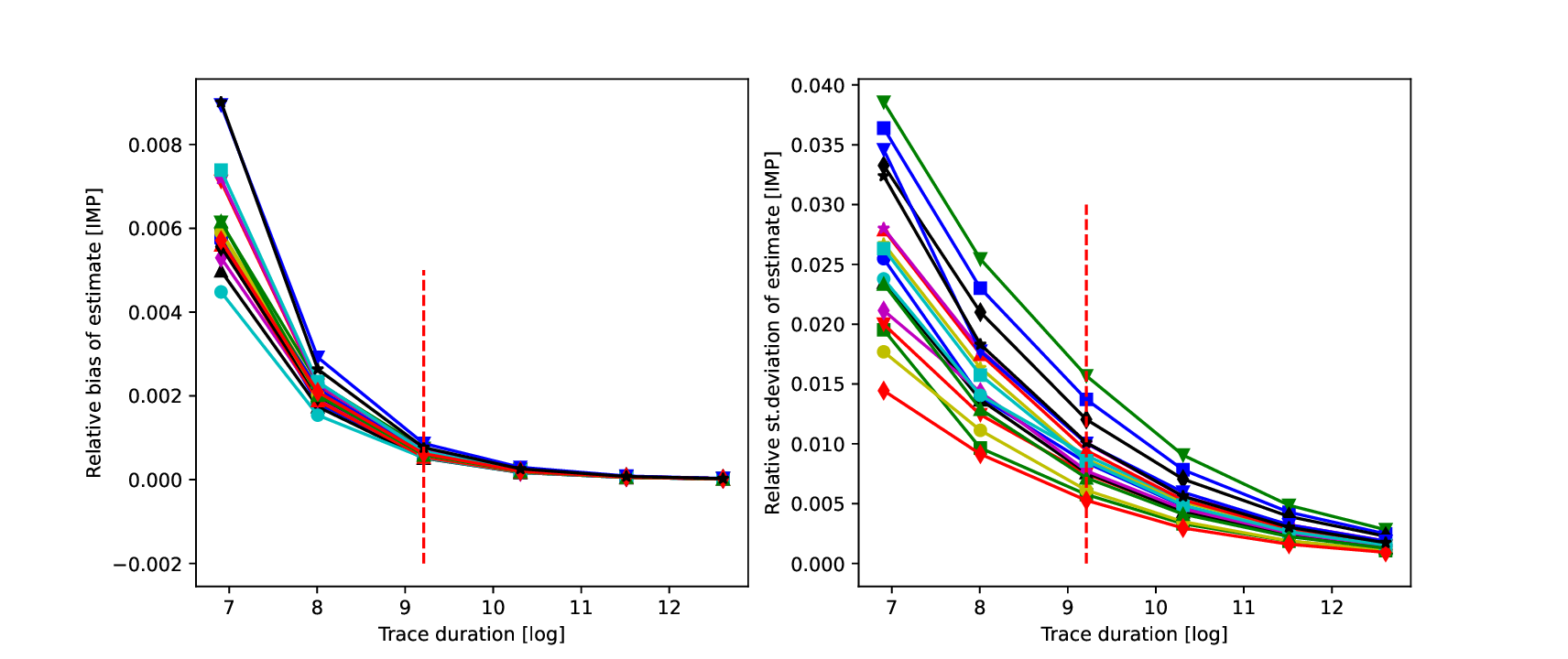}
\caption{Evolution of relative bias and variance in the RL value function (estimates). The vertical dashed line indicates the moment of convergence after which all $bias_{rel}$ and $ \sigma_{rel}$ are below the limits set by the property, and the property is satisfied.}
\centering
\end{figure}

Figure 1 shows the relative bias and standard deviation of Eq.~\ref{property_1} (plotted in logarithmic scale). It can be seen how both bias and standard deviation decay to zero, in agreement with the theory. The vertical dashed line indicates the iteration at which the property is satisfied. It is possible to empirically verify the correctness of the monitor response by confronting the value function obtainable from the actual matrices $T$ and $R$ versus the one that can be computed from estimates $\hat{T}$ and $\hat{R}$ at different iterations of the monitoring. Figure 2 shows the relative error $\frac{V(s)-\hat{V}(s)}{\hat{V}(s)}$ for all 18 states. It can be seen that this error is initially very high, indicating that $\hat{T}$ and $\hat{R}$ are bad estimates. However, the error reduces sensibly with the increase in iterations. At the iteration for which the property is positively validated, it can be seen that the absolute error at such iteration is lower than the estimated bias, confirming the indication of the monitor that both $T$ and $R$ are reasonably learned. 

\begin{figure}[t]
\includegraphics[width=17cm]{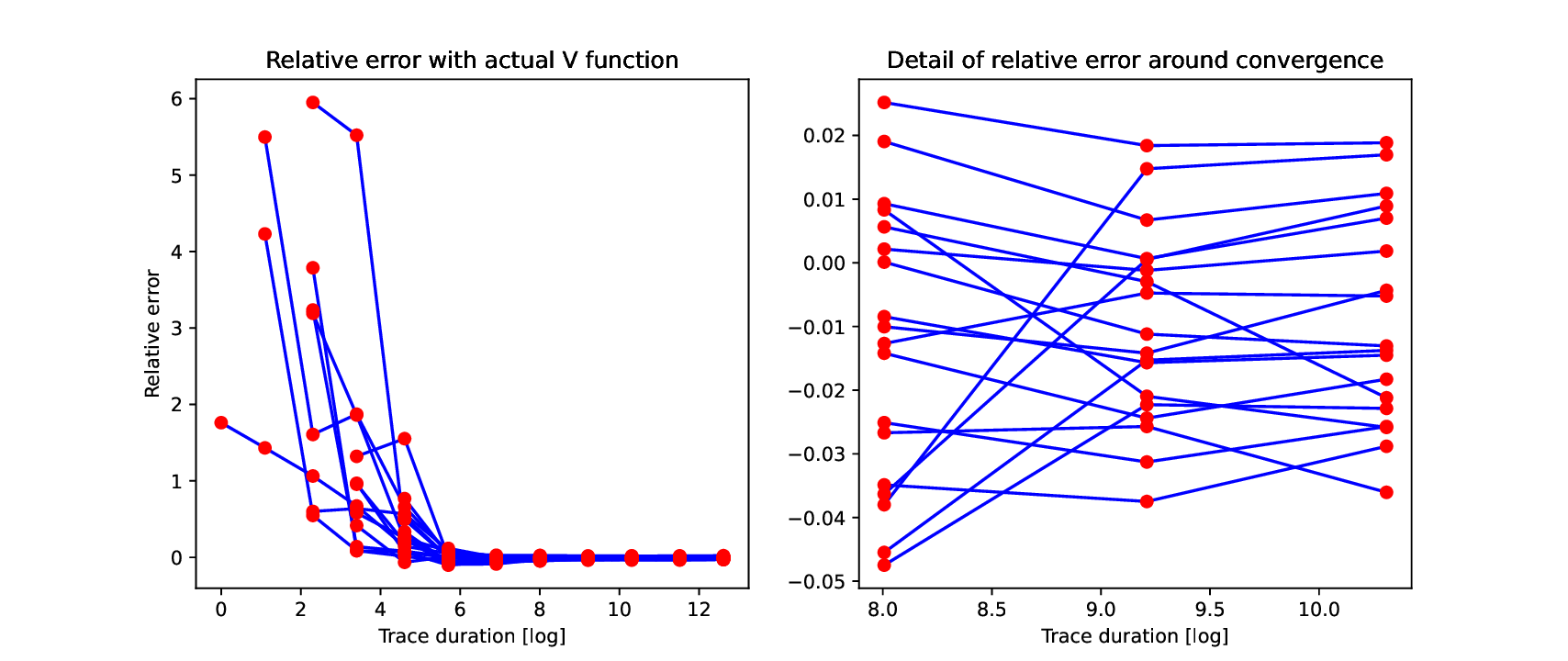}
\caption{Relative error between $V(s)$ and $\hat{V}(s)$ (left); detail when the property is verified (right).  This relative error is an indicator of how well the agent has learned the transition function $T$ and the reward function $R$.}
\centering
\end{figure}

\subsection{How far from the optimum is the current policy?}
\label{relative_quality}

If the optimal value function $V^{\pi^*}$ was known, one could compute how well $\pi$ is faring compared to $\pi^*$. Unfortunately, the optimal value function is not available before learning is concluded. However, it is possible to use the estimate $\hat{V}^{\pi^*}$ as given in Sec.~\ref{fundamentals} in first approximation. 

To simplify the exposition, assume the case of positive definite reward: $R(s, a,s') \geq 0$. In this case, both $\hat{V}^\pi$ and $V^{\pi^*}$ are positive by construction, so that an \textit{optimality ratio} $\eta(s):= \frac{\hat{V}^\pi(s)}{V^{\pi^*}(s)}$ can be defined: if the ratio is sufficiently high for all states, this indicates that the policy is ``almost optimal". 

%

Under the assumption that both value functions $\hat{V}^\pi$ and $\hat{V}^{\pi_\mathrm{cal}} \approx \hat{V}^{\pi^*}$ are normally distributed, it is possible to bound the optimality ratio of $\pi$. Given that the $i^{th}$ diagonal elements $\sigma^2$ of the covariance matrix $cov(\hat{V}^\pi)$ coincide with the variance of the value $\hat{V}^\pi(s_i)$ for the $i^{th}$ state $s_i$,  a 95\% confidence interval in $V^\pi$ can be computed. 

\begin{equation}
    \lfloor{V}^\pi \rfloor := \max(0,\hat{V}^\pi-bias(\hat{V}^\pi)-2\sigma(\hat{V}^\pi)); \: \: \lceil{V}^\pi \rceil := \max(0,\hat{V}^\pi-bias(\hat{V}^\pi)+2\sigma(\hat{V}^\pi)),
\end{equation}

\noindent and similarly for $\hat{V}^{\pi^*}$:

\begin{equation}
    \lfloor{V}^{\pi^*} \rfloor := \max(0,\hat{V}^{\pi^*}-bias(\hat{V}^{\pi^*})-2\sigma(\hat{V}^{\pi^*})); \: \:  \lceil{V}^{\pi^*} \rceil := \max(0,\hat{V}^{\pi^*} )-bias(\hat{V}^{\pi^*})+2\sigma(\hat{V}^{\pi^*})),
\end{equation}

\noindent and thus $\eta$ is bounded as $\underline{\eta} \leq \eta \leq \overline{\eta}$, with

\begin{equation}
\begin{aligned}
    \underline{\eta} =  \lfloor{V}^\pi \rfloor /  \lceil{V}^{\pi^*} \rceil; \: \: 
    \overline{\eta}= \min(1, \lceil{V}^\pi \rceil /  \lfloor{V}^{\pi^*} \rfloor),
\end{aligned}
\end{equation}

\noindent again within a 95\% confidence interval. Omitting the dependency from $s$ for legibility, the procedure is as follows.
\begin{enumerate}
    \item divide the trace into a calibration set and a validation set;
    \item utilize the calibration set to obtain the transition and reward function estimates $\hat{T}_\mathrm{cal}$ and $\hat{R}_\mathrm{cal}$;
    \item obtain the optimal policy $\pi^*_\mathrm{cal}$ via iterated policy improvement;
    \item utilize the validation set to obtain the transition and reward function estimates $\hat{T}_\mathrm{val}$ and $\hat{R}_\mathrm{val}$;
    \item compute the value function $\hat{V}^{\pi_{cal}}$, as well as the bias $bias(\hat{V}^{\pi_{cal}})$ and covariance matrix $cov(\hat{V}^{\pi_{cal}})$ substituting $\hat{T}_\mathrm{val}$ and $\hat{R}_\mathrm{val}$ for $\hat{T}$ and $\hat{R}$;
    \item compute upper and lower bounds $\lfloor{V}^\pi \rfloor$, $\lceil{V}^\pi \rceil$ and $\lfloor{V}^{\pi^*}\rfloor$, $\lceil{V}^{\pi^*}\rceil$, for all states;
    \item compute upper and lower bounds on the optimality ratio $\underline{\eta}$ and $\overline{\eta}$, again for all states;
    \item compare $\underline{\eta}$ and $\overline{\eta}$ with predefined lower thresholds; if for all states the two are above their respective thresholds, the property is satisfied.
\end{enumerate}

\subsubsection*{Example}


 Consider once more the recurring use case example of police patrol. Initially, the system has been provided with a sensible exploratory policy. This is not likely to be the optimal one, but it could be close enough to optimality to not be worth changing. Conversely, it could be so suboptimal to warrant a change of policy. This loose intuition of ``distance from optimum" is encoded in the property:

\begin{equation}
\label{property_2}
    \underline{\eta}(s) \geq 50\% \bigwedge \, \overline{\eta}(s) \geq 70\%  \: \forall s
\end{equation}

%
%
%

\noindent which guarantees that $\pi$ is not too far from the optimum (based on $\underline{\eta}$) as well as indicating that $\pi$ is potentially close to the optimum (based on $\overline{\eta}$). To this must be added the condition

\begin{equation}
\label{condition}
\forall s \max_s \, bias_{rel}(s) < 0.05 \bigwedge  \max_s \, \sigma_{rel}(s) < 0.02, 
\end{equation}

\noindent on both $V^\pi$ and $V^{\pi^*}$, to ensure that all estimates of $T$ and $R$, which are necessary to compute the bounds on $\eta$, are accurate. 

To verify this property (under this condition), it is necessary to first estimate the range of the optimum $V^{\pi^*}$. In this example, the monitor utilizes a random calibration set equal to 5\% of the total traces to estimate the optimum policy. The remainder of 95\% of the traces are utilized to estimate the value function $V^{\pi^*}$ as well as the bias and variance following the procedure introduced in this section. After that, it is possible to compute the optimality ratio boundaries $\underline{\eta}$ and $\overline{\eta}$.

\begin{centering}
\begin{figure}[t]
\includegraphics[width=17cm]{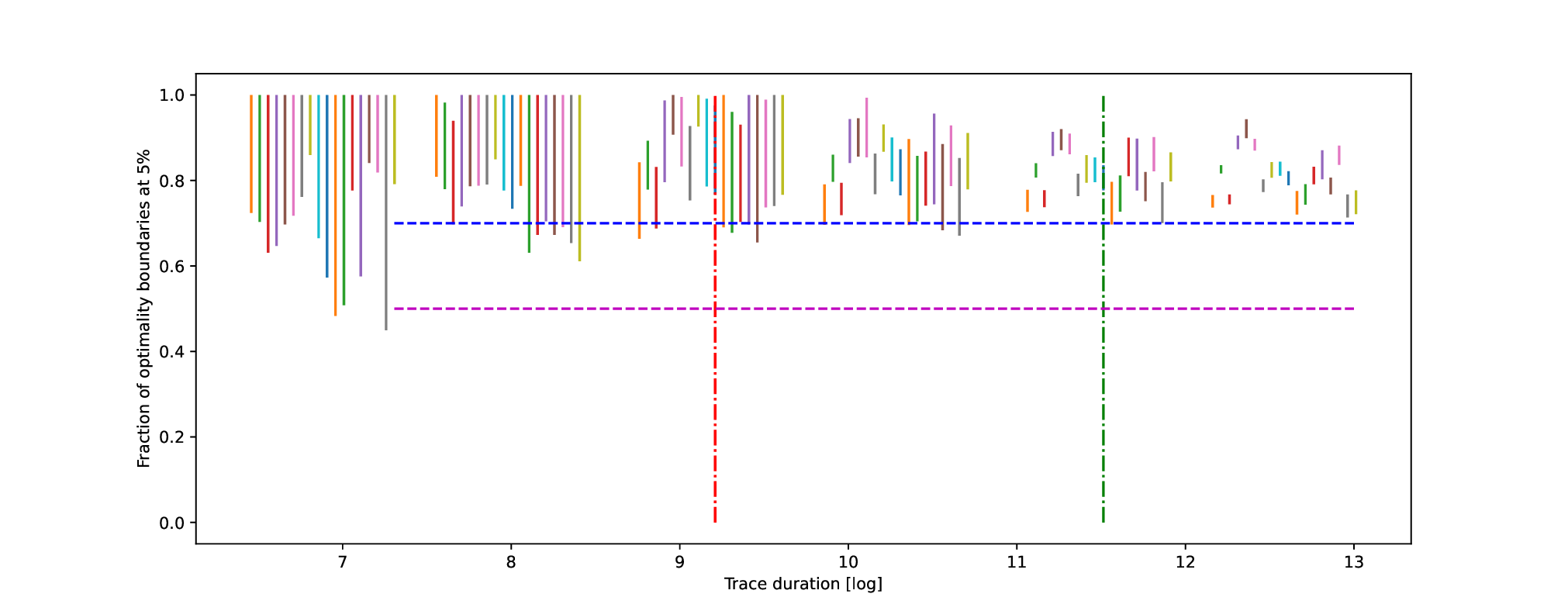}
\caption{Optimality boundaries $\underline{\eta}$ and $\overline{\eta}$, including property thresholds (horizontal dashed lines) and convergence iterations (vertical dot-dashed lines).}
\centering
\end{figure}
\end{centering}

Figure 3 shows the optimality ratio boundaries computed at different times during a sample run of the system. The optimality boundaries are indicated via vertical error bars. The dashed horizontal lines indicate the $50\%$ and $70\%$ optimality thresholds of in Eq.~\ref{property_2}. The vertical lines indicate at which iteration the value function bias and standard deviation satisfy the property in  Eq.~\ref{property_1} for the policy value function $V^\pi$ (left) and for the optimal value function $V^{\pi^*}$ (right) respectively. The example shows that Eq.~\ref{property_2} is punctually verified after  Eq.~\ref{condition} is satisfied.

As can be seen from Figure 3, bounds on $\eta$  shrink with the trace duration; this is due both to the reduction in bias and variance of the estimated optimal value function, as well as due to an actual improvement of the calibrated policy $\pi_{cal}$ due to additional learning. This continuous improvement also explains why error bars at a given iteration are not always contained within the error bars of the previous iterations.

\subsection{How long will it take to evaluate a (new) policy?}
\label{timeliness_property}

We consider three scenarios when an RL system must learn or update a policy:
\begin{enumerate}
    \item the system attempts to introduce a new policy $\pi'$ for the same functionality;
    \item the system functionality change, so that a new unknown reward function $R'$ takes effect;
    \item the system environment changes, so that a new unknown transition function $T'$ takes effect.
\end{enumerate}
Estimating how long this will take might be relevant, especially if there is a finite amount of resources to do so. This section determines an upper bound to the required number of iterations. For all cases, we assume that both policy and learning rate $\alpha$ are stationary. We also assume that the reward function $R$ is bounded and deterministic. To derive this property, we assume an on-policy RL algorithm is used. And our monitor design needs an estimate of the transition function $\hat{T}$.  We discuss this assumption in details when illustrating the third case.   

\subsubsection{New policy}

We analyze first the case when the system desires to implement a different policy, but neither the functionality nor the environment is any different. Estimates $\hat{T}$ and $\hat{R}$ are still valid and refer to a well-learned environment. To predict the learning time, it is necessary to predict how long $V^\pi$ will take to converge to its new value. Consider first the case in which the entire value function $V^\pi$ or action value function $Q^\pi$ can be re-estimated in \textit{updates}, with each update covering the full state space or state-action space, until convergence. This can be done if both $R$ and $T$ are known. In this case, it can be demonstrated that for an on-policy value update, such as a SARSA \cite{Sutton}, the expected consecutive difference $\Delta:= Q^\pi_{k+1} - Q^\pi_k $ before and after an update decays by a factor $\hat{\gamma}=1-\alpha(1-\gamma)$ at each update iteration (see Appendix for details). Furthermore, assuming $R$ is not stochastic, the decay bound holds at every iteration. Defining convergence as $\|\Delta \| < \epsilon$, it can be seen that for an on-policy update, $Q$ will converge in $M_u$ updates, if

\begin{equation}
\label{convergence_bound}
    \hat{\gamma}^{M_u}\|\overline{\Delta}\| < \epsilon \Rightarrow M_u = ceil( \frac{\log(\frac{1}{\epsilon}\|\overline{\Delta}\|)}{\log \frac{1}{\hat{\gamma}}}),
\end{equation}

\noindent where $ceil(x)$ is the ceiling function of $x$, and $\overline{\Delta}$ is an upper estimate of the initial consecutive difference. This is maximum in case $V^\pi_0(s) = Q^\pi_0(s,a) = R_{min}/(1-\gamma)$ and $V^\pi_1(s) = Q^\pi_1(s,a) = R_{max}/(1-\gamma), ~\forall s,a$. Therefore, a rigorous upper bound on convergence can be found when this is re-estimated via full state or state-action updates\footnote{Note that, even though the derivation of $M_u$ makes use of both $T$ and $R$, we do not imply that the RL agent being monitored will make direct use of these quantities.}. 

However, such updates seldom apply in practice, with state visits determined by the stochasticity of $\pi$ and $T$. In this case, it is necessary to reduce the above estimation in terms of $M_u$ updates to a different boundary $M_t$ in terms of \textit{state transitions}. During updates all states or state-action combinations are visited equally often, which is not true for state transitions. However, one could make the rough assumption that if $M_u$ updates are necessary to converge then the same convergence can be reached when each state is visited $M_u$ times (whether or not this assumption is valid will be discussed in due time). Following this first assumption, it is possible to convert the previous bound into

\begin{equation}
\label{first margin}
    M_t = M_u \max_{s,a} \tau(s,a|T,\pi, s_\text{init}),
\end{equation}

\noindent where $\tau(s, a)$ indicates the period, i.e., the number of transitions between two consecutive visits of state $s$ in which action $a$ is selected (also known as the \textit{revisit time}). It is immediate to see that in the most general case, the revisit time is an unbounded stochastic variable; a first approximation of $\tau$ can nonetheless be obtained from the steady-state state probabilities $p$ of the stochastic transition matrix $T$. This can be found as the solution to the following eigenvector problem:

\begin{equation}
\label{steady state prob}
\mathbf{p} :=  \{ p(s_0), ..., p(s_N) \} \Rightarrow \mathbf{p} = T \mathbf{p}, \sum_{i=0}^N p(s_i) = 1, p(s_i)\geq0.
\end{equation}

\noindent Note that to compute such steady-state probability it is necessary to know $T$. That being said, Eq.~\ref{first margin} can be rewritten in first approximation as:

\begin{equation}
\label{second margin}
    M_t = M_u \max_{s,a} [p(s) \pi(s,a)]^{-1}.
\end{equation}

Note that the naive assumption of Eq.~\ref{first margin} that $M_u$ visits at each state-action pair are equivalent to $M_u$ updates does not hold in general. Indeed, updates not only guarantee an equal visit count among all states but also an equal ``mixing'' of visits, so that the entire value or action value function is re-estimated evenly. That being said, in the absence of pathological graphs induced by $T$ (e.g., with absorbing states), one can utilize Eq.~\ref{second margin} as a first upper bound estimate.

Also note that Eq.~\ref{second margin} yields $M_t \rightarrow \infty$ in case either $p(s) \rightarrow 0$ or $\pi(s, a) \rightarrow 0$, i.e., if a state is not reachable or an action for a reachable state is never selected by the policy. In both cases, the limitation is more theoretical than practical. Consider the case $p(\tilde{s})=0$ and assume for simplicity that the environment is not in $\tilde{s}$ at the start of exploration: since $\tilde{s}$ (or $(\tilde{s}, a)$ ) cannot be reached, this never affects the value computation in Eq.~\ref{value_definition} except for $V(\tilde{s})$ (or $Q(\tilde{s}, a)$), and $\tilde{s}$ can be effectively ignored. The same considerations apply for $\pi(s, a)=0$. By extension, a state-action couple for which $p(s) \pi(s, a)$ is smaller than an arbitrarily small negligibility threshold $\omega \geq 0$ will also have a negligible effect on the policy evaluation and can also be ignored. Hence, the new upper bound for convergence in terms of state transitions is  

\begin{equation}
\label{third margin}
    M_t = M_u \max_{(s,a) \in \Omega} [p(s) \pi(s,a)]^{-1}, \: \mathrm{where} \;
    \Omega := \{ (s,a) | p(s) \pi(s,a) \geq \omega \}.
\end{equation}

\noindent Finally, Eq.~\ref{third margin} estimates the transitions required to evaluate the new policy. Selecting a proper $\omega > 0$ is not an immediate choice but requires some insight into the problem at hand. In the absence of this, a conservative estimate can be made by selecting $\omega = 0$, thus ensuring that $M_t$ will be finite. 

Summarizing, the procedure is as follows:

\begin{enumerate}
\item select arbitrarily small coefficients $\epsilon$ and $\omega$;
\item compute bound $M_u$;
\item solve eigenvalue problem of Eq.~\ref{steady state prob} given $T$;
\item compute bound $M_t$ from $M_u$, $\mathbf{p}$ and $\pi$;
\item if $M_t$ is lower than a predefined upper threshold, the property is verified. 
\end{enumerate}

\subsubsection{New reward}
The case of a new reward is identical to the case of a new policy. Indeed, Eq.~\ref{third margin}  can be directly reutilized, with the caveat that $M_u$ must be estimated according to the new values $R_{min}$ and $R_{max}$ if these differ from the previous.

\subsubsection{New environment}
\label{new_environment}
When considering a new environment, $T$ can change from what was previously estimated by the monitor. In principle, this means that the approach investigated in this section is no longer valid due to the unknown state probability $p(s)$. Therefore, the monitor will first need to reestimate $\hat{T}$ (as per the first property discussed). Before that, no estimation can be provided.

That being said, a very conservative, worst-case estimation can still be given if one were to provide a positive negligibility $\omega$ on state-action visits. In that case, a worst-case estimation can be provided in the form:

\begin{equation}
\label{fourth margin}
    \overline{M_t} = M_u \frac{1}{\omega} \geq M_t.
\end{equation}

To apply the previous estimate $M_t$ for runtime monitoring, it is important to make a few observations. First, the convergence and negligibility constants $\epsilon$ and $\omega$ must be provided to the monitor (to use in Eq.~\ref{convergence_bound} and Eq.~\ref{third margin}). Second, the monitor computes the steady-state probability $\mathbf{p}$ of Eq.~\ref{steady state prob} by replacing $T$ with $\hat{T}$. If such an estimate is missing, the monitor can either provide an immediate response using the worst-case estimate of Eq.~\ref{fourth margin} or return an unverified response until the property in Eq.~\ref{property_1} is verified.

\subsubsection*{Example}

Consider once more the police patrol case, and assume that informants notify police authorities of rumoured changes in the location and time of criminal activities. As a result, the initial policy $\pi$ might no longer be satisfactory, due to the unspecified change of the reward function $R$, and reestimating its value will take time. Police officials want to know if the time it will take for the RL system to reestimate correctly the value of $\pi$ is acceptable, i.e., if the following property holds:

\begin{equation}
\label{property_3}
    M_t \leq M^{\text{max}}_t,
\end{equation}

\noindent where $M^{\text{max}}_t$ can be derived from, e.g., a time constraint. A monitor is then deployed to evaluate the property: this must have access to the estimate $\hat{T}$ (since $T$ is unchanged, the estimate is still valid); as well as to the potentially new quantities $R_{min}$ and $R_{max}$. 

Thus, the monitor can employ Eq.~\ref{third margin} to estimate $M_t$ and therefore to verify the property of Eq.~\ref{property_3} before each policy improvement step. As an example, consider the case in which $\pi$ dictates a fixed patrol schedule in the form 

\begin{equation}
    \pi(t, loc) = 
    \begin{cases}
        \text{slums} & \text{if} \: t\leq 1 \\
        \text{station} & \text{if} \: t=2 \\
        \text{docks} & \text{else}.
    \end{cases}
\end{equation}
 
\noindent The monitor can now estimate the number of transitions necessary to evaluate this policy. Given learning parameters $\alpha=0.75$, $\gamma=0.5$, constants $\epsilon=0.05$ and $\omega=0$, and assuming $R_{min}$ and $R_{max}$ to be $0$ and $3$ once more, the monitor estimates 62 transitions necessary for convergence. 
 
Figure 4 shows the actual amount of transitions necessary to estimate the policy value. The continuous lines indicate the change in value function error $\| \Delta \|$ for each of the 18 initial states. It can be seen that $\| \Delta \|$ reduces as the number of transitions increases, albeit not monotonically. It can also be seen that approximately at 30 transitions, the error for all initial states is below the threshold of $\epsilon=0.05$, indicated by the dashed horizontal line. Thus, the value $M_t$ for this example was indeed a conservative estimate. 
 
 \begin{figure}[t]
\includegraphics[width=17cm]{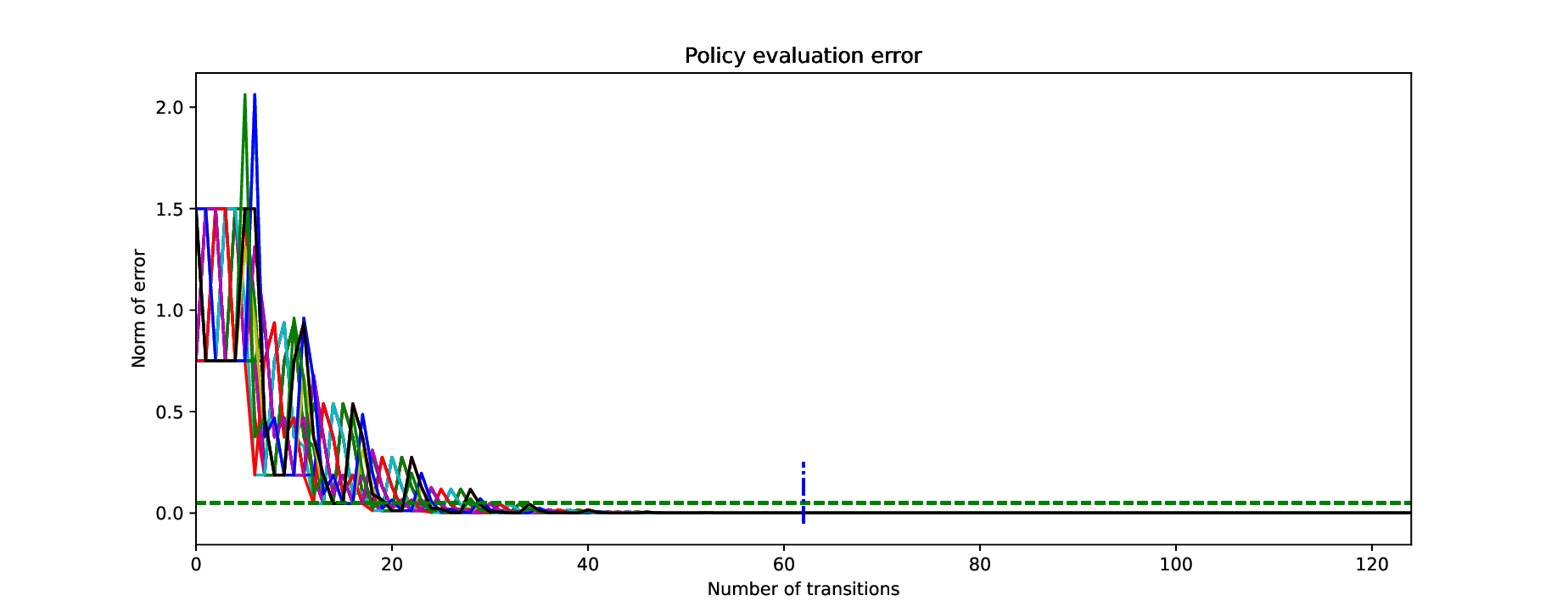}
\caption{Norm of value function error $\| \Delta \|$ for all initial states. The horizontal dotted line indicates the convergence threshold $\epsilon$, while the vertical dot-dashed line indicates the expected convergence time.}
\centering
\end{figure}

\section{Conclusions}
\label{conclusions}
In this paper, we propose three verification properties for the learning phase of reinforcement learning (RL) agents.  The first property relates to the quality of the learning phase.  It expresses whether or not the agent has learned from sufficiently enough and sufficiently varied experiences, to have a suitable representation of its environment.  We devise a runtime verification monitoring technique to assess this property by estimating the variance and bias of the learned value function. This property enables the verification of the second and third properties. 

The second property measures the actual learned policy relative to the ideal optimum.  We extend our monitoring techniques to derive an optimality ratio $\eta$. The verification monitor uses confidence intervals of $\eta$ to indicate upper and lower bounds on optimality. Our RV monitor can check whether the current policy guarantees minimal satisfactory behaviour and whether the policy could potentially be close to optimal.

The third property checks if the learning time falls within a desired number of interactions. It can be used when a new policy is put in place, or when a known policy is applied to a new transition function $T$ or a new reward function $R$. Verification of such property is relevant to systems that require estimating the value induced by a new policy or new environment within a limited time. In this case, the evaluation time is a rough estimate, due to the assumption of Eq.~\ref{first margin} on the equivalence between updates and state transitions. In future work, we will improve this estimate by examining the mixing properties of the graph induced by $\pi$ on the transition matrix $T$.  For this property, we discuss the monitoring techniques necessary to assess the property using the system's runtime observations.    

Our future work includes incorporating these techniques into a runtime verification tool for autonomous robots.  We will also expand the formulation of our properties and monitoring techniques towards Deep Reinforcement Learning algorithms (DRL). DRL algorithms typically use continuous state- and action  spaces, which our approach cannot yet cover. 

\bibliographystyle{eptcs}
\bibliography{FMAS}

\begin{thebibliography}{10}
\providecommand{\bibitemdeclare}[2]{}
\providecommand{\surnamestart}{}
\providecommand{\surnameend}{}
\providecommand{\urlprefix}{Available at }
\providecommand{\url}[1]{\texttt{#1}}
\providecommand{\href}[2]{\texttt{#2}}
\providecommand{\urlalt}[2]{\href{#1}{#2}}
\providecommand{\doi}[1]{doi:\urlalt{https://doi.org/#1}{#1}}
\providecommand{\eprint}[1]{arXiv:\urlalt{https://arxiv.org/abs/#1}{#1}}
\providecommand{\bibinfo}[2]{#2}

\bibitemdeclare{inproceedings}{Anderson}
\bibitem{Anderson}
\bibinfo{author}{Greg \surnamestart Anderson\surnameend},
  \bibinfo{author}{Abhinav \surnamestart Verma\surnameend},
  \bibinfo{author}{Isil \surnamestart Dillig\surnameend} \&
  \bibinfo{author}{Swarat \surnamestart Chaudhuri\surnameend}
  (\bibinfo{year}{2020}): \emph{\bibinfo{title}{Neurosymbolic Reinforcement
  Learning with Formally Verified Exploration}}.
\newblock In: {\slshape \bibinfo{booktitle}{Proceedings of the 34th
  International Conference on Neural Information Processing Systems}},
  \bibinfo{series}{NIPS'20}, \bibinfo{publisher}{Curran Associates Inc.},
  \bibinfo{address}{Red Hook, NY, USA}, \doi{10.48550/arXiv.2009.12612}.

\bibitemdeclare{incollection}{Bartocci}
\bibitem{Bartocci}
\bibinfo{author}{Ezio \surnamestart Bartocci\surnameend},
  \bibinfo{author}{Yli{\`e}s \surnamestart Falcone\surnameend},
  \bibinfo{author}{Adrian \surnamestart Francalanza\surnameend} \&
  \bibinfo{author}{Giles \surnamestart Reger\surnameend}
  (\bibinfo{year}{2018}): \emph{\bibinfo{title}{Introduction to runtime
  verification}}.
\newblock In: {\slshape \bibinfo{booktitle}{Lectures on Runtime Verification}},
  \bibinfo{publisher}{Springer}, pp. \bibinfo{pages}{1--33},
  \doi{10.1007/978-3-319-75632-5}.

\bibitemdeclare{inproceedings}{Corsi}
\bibitem{Corsi}
\bibinfo{author}{Davide \surnamestart Corsi\surnameend},
  \bibinfo{author}{Enrico \surnamestart Marchesini\surnameend} \&
  \bibinfo{author}{Alessandro \surnamestart Farinelli\surnameend}
  (\bibinfo{year}{2021}): \emph{\bibinfo{title}{Formal verification of neural
  networks for safety-critical tasks in deep reinforcement learning}}.
\newblock In: {\slshape \bibinfo{booktitle}{Uncertainty in Artificial
  Intelligence}}, \bibinfo{organization}{PMLR}, pp. \bibinfo{pages}{333--343},
  \doi{10.48448/tj1d-sk77}.

\bibitemdeclare{inproceedings}{Ehlers}
\bibitem{Ehlers}
\bibinfo{author}{Rudiger \surnamestart Ehlers\surnameend}
  (\bibinfo{year}{2017}): \emph{\bibinfo{title}{Formal verification of
  piece-wise linear feed-forward neural networks}}.
\newblock In: {\slshape \bibinfo{booktitle}{International Symposium on
  Automated Technology for Verification and Analysis}},
  \bibinfo{organization}{Springer}, pp. \bibinfo{pages}{269--286},
  \doi{10.1007/978-3-319-68167-2\_19}.

\bibitemdeclare{inproceedings}{Hunt}
\bibitem{Hunt}
\bibinfo{author}{Nathan \surnamestart Hunt\surnameend}, \bibinfo{author}{Nathan
  \surnamestart Fulton\surnameend}, \bibinfo{author}{Sara \surnamestart
  Magliacane\surnameend}, \bibinfo{author}{Trong~Nghia \surnamestart
  Hoang\surnameend}, \bibinfo{author}{Subhro \surnamestart Das\surnameend} \&
  \bibinfo{author}{Armando \surnamestart Solar{-}Lezama\surnameend}
  (\bibinfo{year}{2021}): \emph{\bibinfo{title}{Verifiably safe exploration for
  end-to-end reinforcement learning}}.
\newblock In: {\slshape \bibinfo{booktitle}{{HSCC} '21: 24th {ACM}
  International Conference on Hybrid Systems: Computation and Control,
  Nashville, Tennessee, May 19-21, 2021}}, \bibinfo{publisher}{{ACM}}, pp.
  \bibinfo{pages}{14:1--14:11}, \doi{10.1145/3447928.3456653}.

\bibitemdeclare{article}{Kenton}
\bibitem{Kenton}
\bibinfo{author}{Zachary \surnamestart Kenton\surnameend},
  \bibinfo{author}{Angelos \surnamestart Filos\surnameend},
  \bibinfo{author}{Owain \surnamestart Evans\surnameend} \&
  \bibinfo{author}{Yarin \surnamestart Gal\surnameend} (\bibinfo{year}{2019}):
  \emph{\bibinfo{title}{Generalizing from a few environments in safety-critical
  reinforcement learning}}.
\newblock \doi{10.48550/arXiv.1907.01475}.

\bibitemdeclare{inproceedings}{Li2016}
\bibitem{Li2016}
\bibinfo{author}{Jiwei \surnamestart Li\surnameend}, \bibinfo{author}{Will
  \surnamestart Monroe\surnameend}, \bibinfo{author}{Alan \surnamestart
  Ritter\surnameend}, \bibinfo{author}{Dan \surnamestart Jurafsky\surnameend},
  \bibinfo{author}{Michel \surnamestart Galley\surnameend} \&
  \bibinfo{author}{Jianfeng \surnamestart Gao\surnameend}
  (\bibinfo{year}{2016}): \emph{\bibinfo{title}{Deep Reinforcement Learning for
  Dialogue Generation}}.
\newblock In: {\slshape \bibinfo{booktitle}{Proceedings of the 2016 Conference
  on Empirical Methods in Natural Language Processing}},
  \bibinfo{publisher}{Association for Computational Linguistics},
  \bibinfo{address}{Austin, Texas}, pp. \bibinfo{pages}{1192--1202},
  \doi{10.18653/v1/D16-1127}.
\newblock \urlprefix\url{https://aclanthology.org/D16-1127}.

\bibitemdeclare{article}{Mannor}
\bibitem{Mannor}
\bibinfo{author}{Shie \surnamestart Mannor\surnameend}, \bibinfo{author}{Duncan
  \surnamestart Simester\surnameend}, \bibinfo{author}{Peng \surnamestart
  Sun\surnameend} \& \bibinfo{author}{John~N \surnamestart
  Tsitsiklis\surnameend} (\bibinfo{year}{2007}): \emph{\bibinfo{title}{Bias and
  variance approximation in value function estimates}}.
\newblock {\slshape \bibinfo{journal}{Management Science}}
  \bibinfo{volume}{53}(\bibinfo{number}{2}), pp. \bibinfo{pages}{308--322},
  \doi{10.1287/mnsc.1060.0614}.

\bibitemdeclare{inproceedings}{Mason}
\bibitem{Mason}
\bibinfo{author}{George \surnamestart Mason\surnameend}, \bibinfo{author}{Radu
  \surnamestart Calinescu\surnameend}, \bibinfo{author}{Daniel \surnamestart
  Kudenko\surnameend} \& \bibinfo{author}{Alec \surnamestart Banks\surnameend}
  (\bibinfo{year}{2017}): \emph{\bibinfo{title}{Assured Reinforcement Learning
  with Formally Verified Abstract Policies}}.
\newblock In: {\slshape \bibinfo{booktitle}{Proceedings of the 9th
  International Conference on Agents and Artificial Intelligence, {ICAART}
  2017, Volume 2, Porto, Portugal, February 24-26, 2017}},
  \bibinfo{publisher}{SciTePress}, pp. \bibinfo{pages}{105--117},
  \doi{10.5220/0006156001050117}.

\bibitemdeclare{article}{Mnih}
\bibitem{Mnih}
\bibinfo{author}{Volodymyr \surnamestart Mnih\surnameend},
  \bibinfo{author}{Koray \surnamestart Kavukcuoglu\surnameend},
  \bibinfo{author}{David \surnamestart Silver\surnameend},
  \bibinfo{author}{Alex \surnamestart Graves\surnameend},
  \bibinfo{author}{Ioannis \surnamestart Antonoglou\surnameend},
  \bibinfo{author}{Daan \surnamestart Wierstra\surnameend} \&
  \bibinfo{author}{Martin \surnamestart Riedmiller\surnameend}
  (\bibinfo{year}{2013}): \emph{\bibinfo{title}{Playing atari with deep
  reinforcement learning}}.
\newblock \doi{10.48550/arXiv.1312.5602}.

\bibitemdeclare{article}{Pathak}
\bibitem{Pathak}
\bibinfo{author}{Shashank \surnamestart Pathak\surnameend},
  \bibinfo{author}{Luca \surnamestart Pulina\surnameend} \&
  \bibinfo{author}{Armando \surnamestart Tacchella\surnameend}
  (\bibinfo{year}{2018}): \emph{\bibinfo{title}{Verification and repair of
  control policies for safe reinforcement learning}}.
\newblock {\slshape \bibinfo{journal}{Appl. Intell.}}
  \bibinfo{volume}{48}(\bibinfo{number}{4}), pp. \bibinfo{pages}{886--908},
  \doi{10.1007/s10489-017-0999-8}.

\bibitemdeclare{article}{Potapov}
\bibitem{Potapov}
\bibinfo{author}{Alex \surnamestart Potapov\surnameend} \&
  \bibinfo{author}{MK~\surnamestart Ali\surnameend} (\bibinfo{year}{2003}):
  \emph{\bibinfo{title}{Convergence of reinforcement learning algorithms and
  acceleration of learning}}.
\newblock {\slshape \bibinfo{journal}{Physical Review E}}
  \bibinfo{volume}{67}(\bibinfo{number}{2}), p. \bibinfo{pages}{026706},
  \doi{10.1103/PhysRevE.67.026706}.

\bibitemdeclare{article}{Silver}
\bibitem{Silver}
\bibinfo{author}{David \surnamestart Silver\surnameend},
  \bibinfo{author}{Thomas \surnamestart Hubert\surnameend},
  \bibinfo{author}{Julian \surnamestart Schrittwieser\surnameend},
  \bibinfo{author}{Ioannis \surnamestart Antonoglou\surnameend},
  \bibinfo{author}{Matthew \surnamestart Lai\surnameend},
  \bibinfo{author}{Arthur \surnamestart Guez\surnameend}, \bibinfo{author}{Marc
  \surnamestart Lanctot\surnameend}, \bibinfo{author}{Laurent \surnamestart
  Sifre\surnameend}, \bibinfo{author}{Dharshan \surnamestart
  Kumaran\surnameend}, \bibinfo{author}{Thore \surnamestart
  Graepel\surnameend}, \bibinfo{author}{Timothy~P. \surnamestart
  Lillicrap\surnameend}, \bibinfo{author}{Karen \surnamestart
  Simonyan\surnameend} \& \bibinfo{author}{Demis \surnamestart
  Hassabis\surnameend} (\bibinfo{year}{2017}): \emph{\bibinfo{title}{Mastering
  Chess and Shogi by Self-Play with a General Reinforcement Learning
  Algorithm}}.
\newblock {\slshape \bibinfo{journal}{CoRR}} \bibinfo{volume}{abs/1712.01815},
  \doi{10.48550/arXiv.1712.01815}.

\bibitemdeclare{article}{Singh2021}
\bibitem{Singh2021}
\bibinfo{author}{Bharat \surnamestart Singh\surnameend},
  \bibinfo{author}{Rajesh \surnamestart Kumar\surnameend} \&
  \bibinfo{author}{Vinay~Pratap \surnamestart Singh\surnameend}
  (\bibinfo{year}{2021}): \emph{\bibinfo{title}{Reinforcement learning in
  robotic applications: a comprehensive survey}}.
\newblock {\slshape \bibinfo{journal}{Artificial Intelligence Review}}
  \bibinfo{volume}{55}(\bibinfo{number}{2}), pp. \bibinfo{pages}{945--990},
  \doi{10.1007/s10462-021-09997-9}.

\bibitemdeclare{article}{Singh}
\bibitem{Singh}
\bibinfo{author}{Satinder \surnamestart Singh\surnameend},
  \bibinfo{author}{Tommi~S. \surnamestart Jaakkola\surnameend},
  \bibinfo{author}{Michael~L. \surnamestart Littman\surnameend} \&
  \bibinfo{author}{Csaba \surnamestart Szepesv{\'{a}}ri\surnameend}
  (\bibinfo{year}{2000}): \emph{\bibinfo{title}{Convergence Results for
  Single-Step On-Policy Reinforcement-Learning Algorithms}}.
\newblock {\slshape \bibinfo{journal}{Mach. Learn.}}
  \bibinfo{volume}{38}(\bibinfo{number}{3}), pp. \bibinfo{pages}{287--308},
  \doi{10.1023/A:1007678930559}.

\bibitemdeclare{book}{Sutton}
\bibitem{Sutton}
\bibinfo{author}{Richard~S \surnamestart Sutton\surnameend} \&
  \bibinfo{author}{Andrew~G \surnamestart Barto\surnameend}
  (\bibinfo{year}{2018}): \emph{\bibinfo{title}{Reinforcement learning: An
  introduction}}.
\newblock \bibinfo{publisher}{MIT press}, \doi{10.1109/TNN.1998.712192}.

\bibitemdeclare{inproceedings}{Szepesvari}
\bibitem{Szepesvari}
\bibinfo{author}{Csaba \surnamestart Szepesv{\'{a}}ri\surnameend}
  (\bibinfo{year}{1997}): \emph{\bibinfo{title}{The Asymptotic Convergence-Rate
  of Q-learning}}.
\newblock In: {\slshape \bibinfo{booktitle}{Advances in Neural Information
  Processing Systems 10, {[NIPS} Conference]}}, \bibinfo{publisher}{The {MIT}
  Press}, pp. \bibinfo{pages}{1064--1070}, \doi{10.5555/3008904.3009053}.

\bibitemdeclare{techreport}{Wesel}
\bibitem{Wesel}
\bibinfo{author}{Perry \surnamestart Van~Wesel\surnameend} \&
  \bibinfo{author}{Alwyn~E \surnamestart Goodloe\surnameend}
  (\bibinfo{year}{2017}): \emph{\bibinfo{title}{Challenges in the verification
  of reinforcement learning algorithms}}.
\newblock \bibinfo{type}{Technical Report}.
\newblock \urlprefix\url{https://ntrs.nasa.gov/citations/20170007190}.

\bibitemdeclare{article}{Vinyals}
\bibitem{Vinyals}
\bibinfo{author}{Oriol \surnamestart Vinyals\surnameend}, \bibinfo{author}{Igor
  \surnamestart Babuschkin\surnameend}, \bibinfo{author}{Wojciech~M
  \surnamestart Czarnecki\surnameend}, \bibinfo{author}{Micha{\"e}l
  \surnamestart Mathieu\surnameend}, \bibinfo{author}{Andrew \surnamestart
  Dudzik\surnameend}, \bibinfo{author}{Junyoung \surnamestart
  Chung\surnameend}, \bibinfo{author}{David~H \surnamestart Choi\surnameend},
  \bibinfo{author}{Richard \surnamestart Powell\surnameend},
  \bibinfo{author}{Timo \surnamestart Ewalds\surnameend},
  \bibinfo{author}{Petko \surnamestart Georgiev\surnameend} et~al.
  (\bibinfo{year}{2019}): \emph{\bibinfo{title}{Grandmaster level in StarCraft
  II using multi-agent reinforcement learning}}.
\newblock {\slshape \bibinfo{journal}{Nature}}
  \bibinfo{volume}{575}(\bibinfo{number}{7782}), pp. \bibinfo{pages}{350--354},
  \doi{10.1038/s41586-019-1724-z}.

\bibitemdeclare{article}{Watkins1992}
\bibitem{Watkins1992}
\bibinfo{author}{Christopher~J.C.H. \surnamestart Watkins\surnameend} \&
  \bibinfo{author}{Peter \surnamestart Dayan\surnameend}
  (\bibinfo{year}{1992}): {\slshape \bibinfo{journal}{Machine Learning}}
  \bibinfo{volume}{8}(\bibinfo{number}{3/4}), pp. \bibinfo{pages}{279--292},
  \doi{10.1023/a:1022676722315}.

\bibitemdeclare{article}{Xin}
\bibitem{Xin}
\bibinfo{author}{Bo~\surnamestart Xin\surnameend}, \bibinfo{author}{Haixu
  \surnamestart Yu\surnameend}, \bibinfo{author}{You \surnamestart
  Qin\surnameend}, \bibinfo{author}{Qing \surnamestart Tang\surnameend} \&
  \bibinfo{author}{Zhangqing \surnamestart Zhu\surnameend}
  (\bibinfo{year}{2020}): \emph{\bibinfo{title}{Exploration entropy for
  reinforcement learning}}.
\newblock {\slshape \bibinfo{journal}{Mathematical Problems in Engineering}}
  \bibinfo{volume}{2020}, \doi{10.1155/2020/2672537}.

\bibitemdeclare{inproceedings}{Zhu}
\bibitem{Zhu}
\bibinfo{author}{He~\surnamestart Zhu\surnameend}, \bibinfo{author}{Zikang
  \surnamestart Xiong\surnameend}, \bibinfo{author}{Stephen \surnamestart
  Magill\surnameend} \& \bibinfo{author}{Suresh \surnamestart
  Jagannathan\surnameend} (\bibinfo{year}{2019}): \emph{\bibinfo{title}{An
  inductive synthesis framework for verifiable reinforcement learning}}.
\newblock In: {\slshape \bibinfo{booktitle}{Proceedings of the 40th {ACM}
  {SIGPLAN} Conference on Programming Language Design and Implementation,
  {PLDI} 2019, Phoenix, AZ, USA, June 22-26, 2019}},
  \bibinfo{publisher}{{ACM}}, pp. \bibinfo{pages}{686--701},
  \doi{10.1145/3314221.3314638}.

\end{thebibliography}

\section*{Appendix}
First, observe that the action value update can be rewritten as

\begin{equation*}
\label{rewritten_1}
    Q^\pi_{k+1}(s,a) = (1-\alpha)Q^\pi_k(s,a)+ \alpha \sum_{s' \in S} T(s,a,s') [r(s,a, s')+\gamma \sum_{a'} \pi(s',a') Q^\pi_k(s', a)].
\end{equation*}

\noindent Define now the matrix $\Pi \in \mathbb{R}^{|S|,|A|,|S|}$, whose entries are defined as:
\begin{equation*}
\Pi_{i,j} = 
\begin{cases}
\pi(s_j, a_{i-(j-1)|A|}) & \text{if} (j-1)|S||A| < i \leq j|S||A| \\
0 & \text{otherwise}.
\end{cases}
\end{equation*}
 
\noindent $\Pi$ is the matrix obtained by staking $|S|$ copies of the vector-wise policy $\pi \in \mathbb{R}^{|S|,|A|}$, and setting to zero each element of the resulting $j^\text{th}$ column which does not contain the probabilities for the $j^\text{th}$ state. Denote as $\mathbf{Q} \in \mathbb{R}^{|S|,|A|}$ the column vector of values $Q^\pi(s,a)$, as $\mathbf{R} \in \mathbb{R}^{|S|,|A|}$ the column vector of the expected rewards $E_{s' \in S}[r(s,a,s')]$, and as $\mathbf{T} \in \mathbb{R}^{|S|,|A|,|S|}$ the matrix corresponding to the transition function $T$. Then the value update can be written in matrix form as

 

 \begin{equation*}
 \label{matricial_form}
    \mathbf{Q}_{k+1} = \alpha \mathbf{R} + [\alpha  \gamma \mathbf{T} \Pi^T  + (1-\alpha)\mathbf{I}] \mathbf{Q}_k = \alpha \mathbf{R} + \mathbf{B} \mathbf{Q}_k,
\end{equation*}
 
\noindent where $\mathbf{I}$ is the identity matrix. Note now that the square matrix $\mathbf{B} := [\alpha  \gamma \mathbf{T} \Pi^T  + (1-\alpha)\mathbf{I}]$ has norm $\| \mathbf{B} \| \leq \alpha \gamma + 1-\alpha = 1-\alpha(1-\gamma)$ by construction, because $\alpha, \gamma \leq 1$, and $\mathbf{T}, \Pi$ are stochastic matrices. 

If we define the consecutive error vector as $\Delta_k := \mathbf{Q}_{k+1} - \mathbf{Q}_k$. Then it follows that $\Delta_{k+1} = \mathbf{B} \Delta_k$ indicates a contraction since $1-\alpha(1-\gamma) \leq 1$.

\end{document}